\documentclass{article}

     \PassOptionsToPackage{numbers, compress}{natbib}

\usepackage[preprint]{neurips_2021}
\usepackage{times}
\usepackage{epsfig}
\usepackage{graphicx}
\usepackage{amsmath,amsthm}
\usepackage{amssymb}
\usepackage{pifont}
\usepackage{dsfont}
\usepackage{textcomp}
\usepackage{wrapfig}
\usepackage{multirow}
\usepackage{overpic}
\usepackage[export]{adjustbox}
\usepackage{tikz,pgfplots}
\usepackage[toc]{appendix}

\usepackage[pagebackref=true,breaklinks=true,letterpaper=true,colorlinks,bookmarks=false]{hyperref}

\usepackage{subcaption}
\usepackage{cleveref}
\usepackage{xcolor}

\newcommand{\net}{\mathcal{G}}
\newcommand{\loss}{\mathcal{L}}
\newcommand{\netTwo}{\mathcal{G}_1}
\newcommand{\netThree}{\mathcal{G}_2}
\newcommand{\R}{\mathbb{R}}
\newcommand{\inputSpace}{\mathcal{X}}
\newcommand{\outputSpace}{\mathcal{Y}}
\newcommand{\invarianceSet}{S}
\newcommand{\comment}[1]{}
\newcommand{\orbitMapping}{h}
\DeclareMathOperator{\argmax}{\arg \max}

\newtheorem{proposition}{Proposition}
\newtheorem{fact}{Fact}
\crefname{fact}{Fact}{Facts}
\crefname{proposition}{Proposition}{Propositions}
\newtheorem{example}{Example}
\crefname{example}{Example}{Examples}

\begin{document}

\title{Training or Architecture? How to Incorporate Invariance in Neural Networks}
%
%
%
%
\author{Kanchana Vaishnavi Gandikota, Jonas Geiping, Zorah L\"ahner,
Adam Czapli\'{n}ski, Michael Moeller\\
University of Siegen\\
{\tt\small first name (dot) last name @ uni-siegen.de}
}

\maketitle

\begin{abstract}

Many applications require the robustness, or ideally the invariance, of a neural  
network to certain transformations of input data. Most commonly, this requirement is addressed by either augmenting the training data, using adversarial training, or defining network architectures that include
 the desired invariance automatically. Unfortunately, the latter often relies on the ability to enlist all possible transformations, which make such approaches largely infeasible for infinite sets of transformations, such as arbitrary rotations or scaling. 
 In this work, we propose a method for provably invariant network architectures with respect to group actions by choosing one element from a (possibly continuous) orbit based on a fixed criterion.  
 In a nutshell, we intend to 'undo' any possible transformation before feeding the data into the actual network. We analyze properties of such approaches, extend them to equivariant networks, and demonstrate their advantages in terms of robustness as well as computational efficiency in several numerical examples. In particular, we investigate the robustness with respect to rotations of images (which can possibly hold up to discretization artifacts only) as well as the provable rotational and scaling invariance of 3D point cloud classification. 
\end{abstract}

\section{Introduction}
Deep neural networks have revolutionized the field of computer vision over the past decade. Yet, training deep networks in a straight-forward way often leads to a lack of robustness and prevents a safe usage of the resulting predictors. In image classification, for instance, rotational, scale, and certain types of shift invariance are often highly desirable properties. Despite the use of millions of realistic images in datasets like Imagenet \cite{imagenet}, networks trained on such data do not inherit desirable properties automatically. On the contrary, they seem to be susceptible to adversarial attacks with respect to these transformations (see e.g. \cite{engstrom2017rotation,finlayson2019adversarial}), and small perturbations can cause significant changes in the networks' predictions. To counteract this behavior, the two major directions of research are to either modify the training procedure or the network architecture, the advantages and drawbacks of which are the main focus of this work. 

\begin{figure}
\centering

\begin{subfigure}[b]{.26\textwidth}
\centering
\includegraphics[width=.45\textwidth]{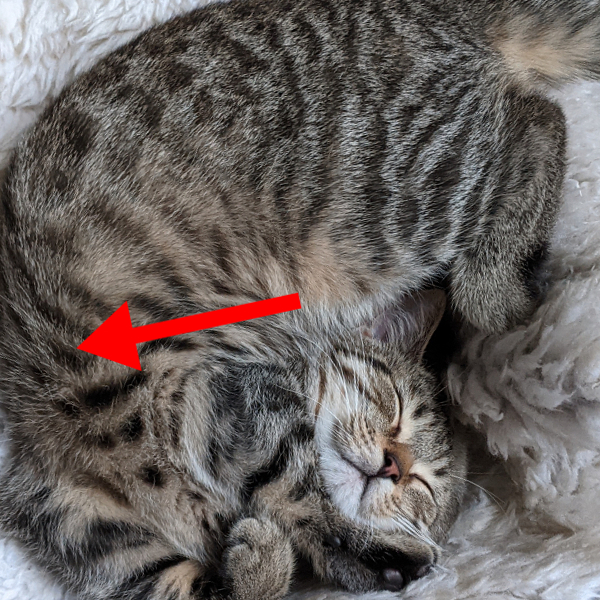}\enskip
\includegraphics[width=.45\textwidth]{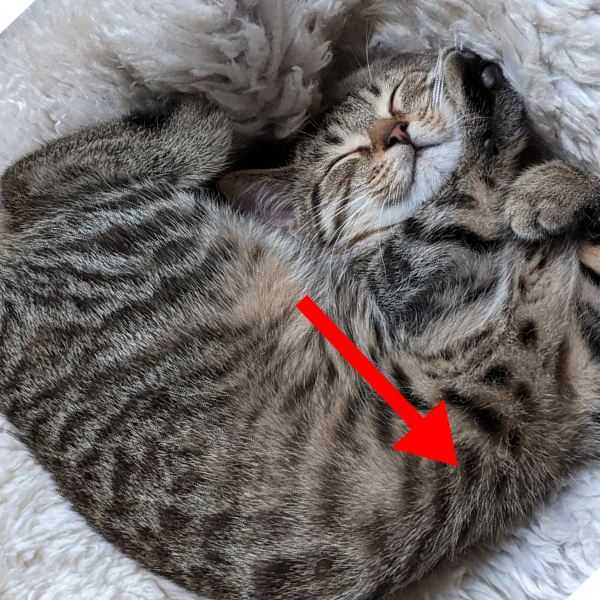}\\
\includegraphics[width=.45\textwidth]{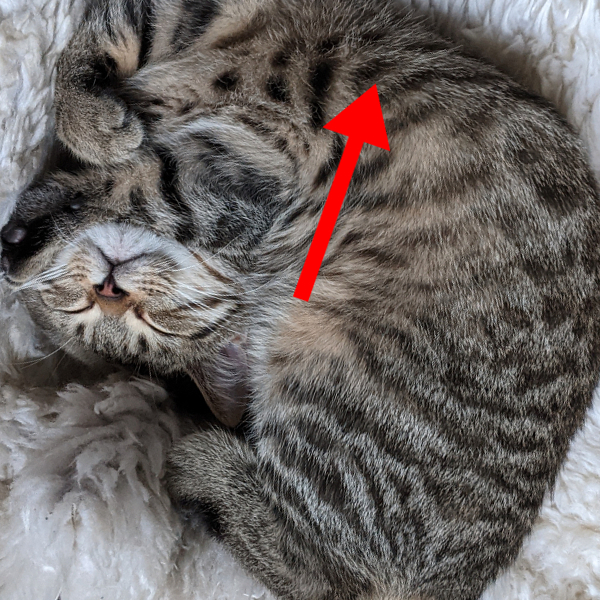}\enskip
\includegraphics[width=.45\textwidth]{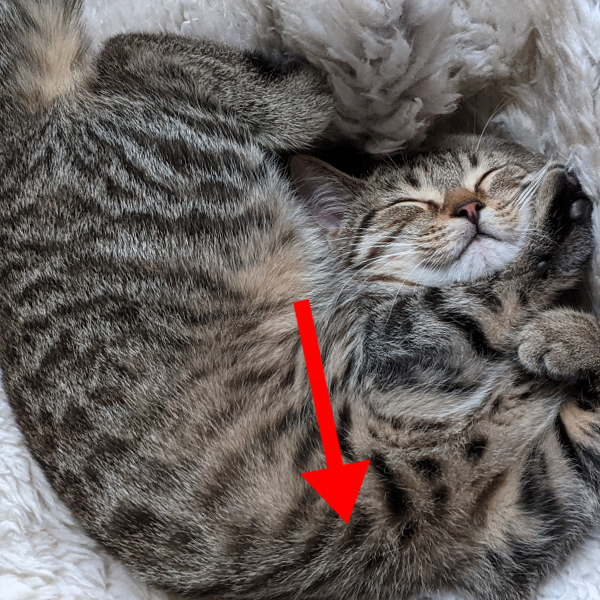}\quad
\caption{Samples of the orbit}
\end{subfigure}
\begin{subfigure}[b]{.24\textwidth}
\centering
\includegraphics[width=\textwidth]{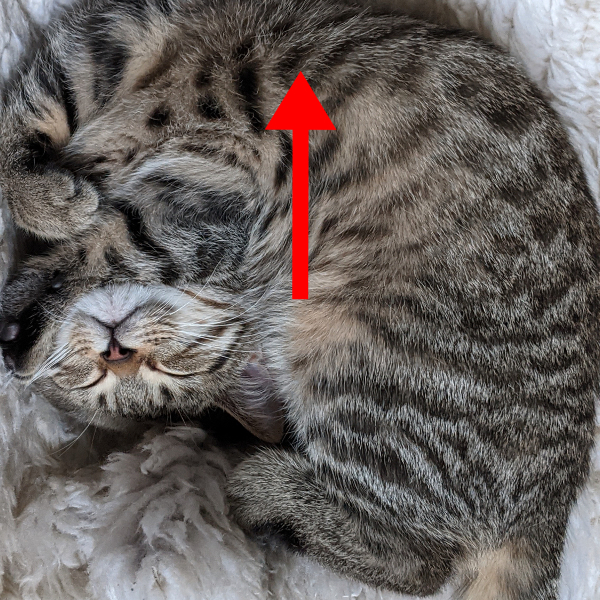}\quad
\caption{Orbit mapping element}
\end{subfigure}
\begin{subfigure}[c]{.4\textwidth}
\centering
\includegraphics[width=\textwidth,margin={0 0 0 -4cm}]{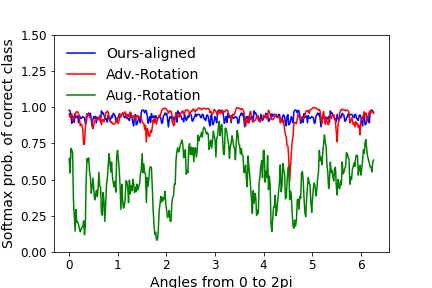}
\caption{}
\end{subfigure}
    \caption{(Left) Picture of a cat in four different rotations which are four samples from the orbit of all rotated versions of this image. The mean gradient direction is marked in red. 
    We choose the element with the mean gradient pointing upwards for the orbit mapping.
    (Right) The plot shows accuracy when rotating a sample image by $0$ to $360$ degrees and the softmax probabilities of the true label are reported for three different networks over all rotations. Clearly, data augmentation is not sufficient for a guaranteed robustness of the classifier w.r.t. continuous rotations. }
    \label{fig:teaser}
\end{figure}

Modifications of the training procedure replace the common training of a network $\net$ with parameters $\theta$ on training examples $(x^i,y^i)$ via a loss function $\loss$,
\begin{align}
    \min_\theta \sum_{\text{examples i}} \loss(\net(x^i;\theta);y^i),
\end{align}
with a loss function that considers all perturbations in a given set $\invarianceSet$ of transformations to be invariant towards. The most common choices are taking the mean loss of all predictions $\{\net(g(x^i);\theta) ~|~ g \in \invarianceSet\}$ (resulting in training with \textit{data augmentation}), or the maximum loss among all predictions (resulting in \textit{adversarial training}). Unfortunately, such training schemes can never yield provable invariance. In particular, training with data augmentation is far from yielding robust results with respect to transformations as illustrated in  Fig.~\ref{fig:teaser} at the example of rotations: The plot shows the softmax probabilities of the true label when feeding the exemplary image at various rotations ranging from 0 to $2\pi$ into a network trained with rotational augmentation (green), and adversarial training (red). As we can see, rotational data augmentation is not nearly sufficient to truly make a classification network robust towards rotations, and even the significantly more expensive adversarial training shows instabilities. 

While modifications of the training scheme remain the best generic option (particularly for complex or hard-to-characterize transformations), more structured transformations, e.g., those arising from a group action, allow to modify the network architecture such that it yields a provable invariance. As opposed to previous works that largely rely on the ability to enlist all transformations of an input $x$ (i.e., assume a finite \textit{orbit}), we propose to make network architectures invariant by selecting a specific element from a (possibly infinite) orbit generated by a group action, through an application-specific mapping. Simply put, we are trying to undo and fix the transformation/pose of the input data \textit{before} feeding it into the main network. We subsequently generalize the theory of provable invariance to provable \textit{equivariance}, i.e., architectures that commute with a set of certain transformations, $\net(g(x);\theta) = g(\net(x;\theta))$ for all $g \in \invarianceSet$. 

Our proposed approach is significantly easier to train than adversarial training methods while being at least equally performant and robust, as well as computationally cheaper.
We illustrate these findings on the rotation invariant classification of images (on which discretization artifacts from the interpolation after any rotation play a crucial role) as well as on the scale, rotation, and translation invariant classification of 3D point clouds (for which the invariance can be guaranteed).

Our contributions can be summarized as follows:
\begin{itemize}
    \item We propose a general way of designing network architectures that are in- (or equi-)variant to transformations from sets $\invarianceSet$ that are associated with a group action.
    \item We demonstrate that a combination of data augmentation with the proposed orbit mapping is a simple way for provable invariance with almost no computational overhead (in contrast to adversarial training). 
    \item We obtain state-of-the-art results for the worst-case classification accuracy over all possible rotations of images, and demonstrate that orbit mappings extend to other applications such as 3D point cloud classification. 
\end{itemize}
\section{Related Work}
\label{sec:related}
Several approaches have been developed in the literature to encourage models to exhibit invariance or robustness to desired transformations of data. These include i)~data augmentation using desired transformations, ii)~use of regularization to encourage the network output to remain constant for desired transformations on the input \cite{simard1991tangent}, iii) use of adversarial training~\cite{engstrom2019exploring} and regularization~\cite{yang2019invariance}, iv)~use of hand-crafted invariant features for downstream classification tasks \cite{sheng1994orthogonal,yap2010polar,tan1998rotation,MANTHALKAR20032455}, v)~incorporating desired invariance properties in to the neural network design~\cite{cohen2016group, Worrall_2017_CVPR,Weiler2019GeneralES}, and vi) train time/test time data transformation. The approaches i)-iii) can lead to improved robustness but cannot yield provable invariance to transformations. Hand-crafting features can yield desired invariance, but is difficult and often sacrifices accuracy. Provable invariance to a finite number of transformations can be achieved by applying all such transformations to the each input data point and pooling the corresponding features  \cite{manay2006integral, Reisert2008,laptev2016ti}. This strategy can even be applied only during test time. It can, however, not be extended to sets with infinitely many transformations such as arbitrary rotations.

Recent approaches \cite{cohen2016group, ravanbakhsh2017equivariance,Weiler2019GeneralES} incorporate in-/equivariances when the desired transformations of the data can be formulated as a group action, e.g. guaranteeing equivariance but enforcing this property in each layer separately. 
Layer wise approaches for equivariance to finite groups such as \cite{cohen2016group} typically use all possible transformations at each layer. 
In contrast to layer-wise approaches, our approach is to  enforce in- and equivariance by consistently selecting a specific element in the orbit. 
A  similar approach is also adopted by \cite{esteves2018polar, tai2019equivariant}, which uses an application-dependent coordinate transformation and a spatial transformer network \cite{jaderberg2015transformer}  to  learn to undo transformations. Learning to undo a transformation, however, requires additional training with data augmentation and cannot yield provable invariance. In contrast, we propose to directly select an element from the orbit of the input data, which can improve robustness without data augmentation, and can even be applied at test time only.

As we will show experiments for rotationally invariant image classification, as well as for orientation and permutation invariant 3D object classification in the later part of this paper, let us summarize specific prior work related to in- and equivariance in these application.

\paragraph{Provable Rotational In-/equivariance in 2D}
Several works \cite{cohen2016group, marcos2017rotation, veeling2018rotation, marcos2016learning} have considered layer wise equivariance to discrete rotations using multiple rotated versions of filters at each layer, which was formalized using group convolutions in \cite{cohen2016group}.  For equivariance to continuous rotations Worrall et al. \cite{Worrall_2017_CVPR} utilize circular harmonic filters at each layer. All these layer wise approaches for group equivariance in images were unified in a single framework in \cite{Weiler2019GeneralES}. Instead of layer-wise approaches, \cite{fasel2006rotation, laptev2016ti, henriques2017warped} input multiple rotated copies of images to the network and pool the corresponding features. Most closely related to our approach are works that align images in a specific direction, for example using a principal component analysis (PCA) for binary images \cite{rehman2018automatic} (which, however, does not extend to grayscale images), a radon transform \cite{1424459} (for which a fine discretization of continuous rotations is expensive), or using a learnable transformation \cite{jaderberg2015transformer, tai2019equivariant} (which cannot guarantee the invariance).

\paragraph{Rotation Invariance in 3D}
Due to the different representations of 3D data (e.g. voxels, point clouds, meshes), many different strategies exist. On voxel representations, some 2D imaging techniques can be generalized and adapted, e.g. probing several rotations during test time \cite{wu2015shapenets, Wang-2017-ocnn}, or creating rotationally equivariant kernels for convolutions \cite{weiler20183d,thomas2018tensor}. 
\cite{esteves2018polar} proposes a polar transformer network, that works for both images and voxel grids, learning a transformation into a log-polar space in which rotations become translations such that they can be handled via usual convolutions.

Point cloud representations remove problems like discretization artifacts after rotations, but struggle with less clear neighborhood information due to unordered coordinate lists. 
\cite{ZhangR_18_gcnn_point_cloud} solve this by adding hierarchical graph connections to the point clouds and using graph convolutions. While graph convolutions are independent of rotations, the learned features still depend on the rotation of the input data. 
The classical PointNet architecture \cite{qi2017pointnet} contains a spatial transformer network \cite{jaderberg2015transformer} as a built-in pre-processing step, and uses data augmentation in the hope to learn a proper invariance.  
Its extension, PointNet++ \cite{qi2017pointnetpp}, additionally considers hierarchical and neighborhood information. Both approaches, however, cannot guarantee an invariance.

On the other hand, local operations and convolutions on the surface of triangular meshes are invariant to global rotations by definition \cite{monti2016monet} but, due to their local nature, do not capture global information. 
MeshCNN \cite{hanocka2019meshcnn} addresses this by adding pooling operations through edge collapse. 
\cite{sharp2020diffusion} defines a representation independent network structure based on heat diffusion which can balance between local and global information.
\cite{deng2018ppf} and \cite{zhao20193d} achieve rotation invariance on point clouds by considering pairs of features in the tangent plane of each point. 

\comment{\subsection{Scale invariance/equivariance}
\cite{MANTHALKAR20032455} make use of wavelet filter banks to extract features invariant to rotation and scaling transformations.\\
\cite{kanazawa2014locally}-employ multiple scaled versions of convolution filters  at each layer, and perform max pooling  of filter responses over scale. (similar to using rotated filters- experiments on scaled MNIST. They measure invariance using method proposed in \cite{goodfellow2009measuring})\\
\cite{NEURIPS19_scale} - scales pace as a semigroup, they derive group convolution by truncating the scale-space to finite integer down-scale factors,  (somewhat similar to group convolution paper-show results on PCAM and cityscapes)\\
\cite{Sosnovik2020Scale-Equivariant} scale-equivariant convolutional networks with steerable filters
\subsection{combined equivariance}
\cite{esteves2018polar} polar transformer networks invariant to translation and equivariant to rotation and scaling transformations, by making use of  a differentiable log-polar transformation module, and training an additional network to predict the origin in polar coordinates (experiments on rotated mnist and 3d object classification). }

\section{Proposed Approach}
\label{sec:proposedApproach}

\subsection{In- and Equivariant Networks w.r.t. Group Actions}

We consider a network $\net$ to be a function $\net: \inputSpace \times \R^p \rightarrow \outputSpace$ that maps data $x \in \inputSpace$ from some suitable input space $\inputSpace$ to some prediction $\net(x;\theta) \in \outputSpace$ in an output space $\outputSpace$, where the way this mapping is performed depends on parameters $\theta \in \R^p$. We wonder how, for a given set $\invarianceSet \subset \{g: \inputSpace \rightarrow \inputSpace\}$ of transformations of the input data, we can achieve the \textit{invariance} of $\net$ to $\invarianceSet$ defined as
\begin{align}
\label{eq:invariance}
    \net(g(x);\theta) = \net(x;\theta) \quad \forall x \in \inputSpace, ~ g \in \invarianceSet, ~ \theta \in \R^p.
\end{align}

Closely related, is the concept of  \textit{equivariance} of $\net$ which preserves the structure  of transformations $g\in \invarianceSet$ of input data in the elements $y \in \outputSpace$ (including, but not limited to, the case where $\inputSpace \equiv \outputSpace$). The \textit{equivariance} of $\net$ to $\invarianceSet$ is defined as
\begin{align}
\label{eq:equivariance}
    \net(g(x);\theta) = g(\net(x;\theta)) \quad \forall x \in \inputSpace, ~ g \in \invarianceSet, ~ \theta \in \R^p.
\end{align}
The in- and equivariance of a network with respect to transformations in $\invarianceSet$ is of particular interest in the case where $\invarianceSet$ induces a \textit{group action}\footnote{A (left) group action of  a group $S$ with the identity element $e$,  on a set $X$ is a map 
$\sigma : S\times X\rightarrow X,$
that satisfies i)~$\sigma(e,x)=x$ and  ii)~$\sigma(g,\sigma(h,x))=\sigma(gh,x)$,  $\forall g,h\in S$ and  $\forall x\in X$.  When the action being considered is clear from the context, we write  $g(x)$ instead of $\sigma(g,x)$.
} on $\inputSpace$, which is what we will assume about $\invarianceSet$ for the remainder of this paper. 
Of particular importance for the construction of both, invariant and equivariant networks, is the set of all possible transformations of input data $x$,
\begin{align}
\label{eq:orbit}
\invarianceSet \cdot x = \{g(x) ~|~ g \in \invarianceSet\},
\end{align}
which is called the \textit{orbit of $x$}. A basic observation for constructing invariant networks is that any network acting on the orbit of the input is automatically invariant to transformations in $\invarianceSet$: 
\begin{fact}\textbf{Characterization of Invariant Functions via the Orbit:}
\label{fact:invariance}
Let $\invarianceSet$ define a group action on $\inputSpace$. A network $\net: \inputSpace \times \R^p \rightarrow \outputSpace$ is invariant under the group action of $\invarianceSet$ if and only if it can be written as $\net(x;\theta) =  \netTwo(\invarianceSet \cdot x;\theta)$ for some other 
network $\netTwo: 2^\inputSpace \times \R^p \rightarrow \outputSpace$. 
\end{fact}
The above observation is based on the fact that $\invarianceSet \cdot x = \invarianceSet \cdot g(x)$ holds for any $g \in \invarianceSet$, provided that $\invarianceSet$ is a group. Although not taking the general perspective of \cref{fact:invariance}, the approaches \cite{ Reisert2008, laptev2016ti}, which integrate (sum over finite elements) the mappings of $\net$ over a  (discrete) group, can be interpreted as instances of \cref{fact:invariance}, where  $\netTwo$ corresponds to the  summation. 

Similar strategies of applying all transformations in $\invarianceSet$ to the input $x$ can be pursued for the design of equivariant layers:
\begin{proposition}
\label{fact:equivariance}
Let $\invarianceSet$ define a group action on $\inputSpace$. A network $\net$ is equivariant under the group action of $\invarianceSet$ if it can be written as 
\begin{align}
    \label{eq:fact2}
\net(x;\theta) =  \netTwo(\{g (\netThree (g^{-1}(x);\theta_2)) ~|~ g \in \invarianceSet\};\theta_1)
\end{align}
for some other arbitrary network $\netThree: \inputSpace \times \R^{p_2} \rightarrow \inputSpace$, and a network $\netTwo: 2^\inputSpace \times \R^{p_1} \rightarrow \inputSpace$ that commutes with any element $h \in \invarianceSet$, i.e., for $h \in \invarianceSet$, and $Z\subset \inputSpace$, it satisfies
$\netTwo(h(Z);\theta_2) = h(\netTwo(Z;\theta_2))$, where $h(Z)$ denotes the set obtained by the applying $h$ to every element of $Z$. 
\end{proposition}
We refer to the appendix for a proof. The work \cite{cohen2016group} can be interpreted as an instance of the construction in \cref{fact:equivariance}, where equivariant linear layers w.r.t. rotations by 90 degrees are obtained by choosing $\netThree$ to be a simple convolution and $\netTwo$ to be the summation over all (finitely many) elements of the set. Subsequently, they nest these layers with component-wise (and therefore inherently equivariant) non-linearities.

\subsection{Orbit Mappings}
While \cref{fact:invariance,fact:equivariance} are stated for general (even infinite) groups, realizations of these constructions from the literature often rely on the ability to list an entire orbit of the group. In this work we would like to include an efficient solution even for cases in which the orbit is not finite, and utilize \cref{fact:invariance} in the most straight-forward way: We propose to construct provably invariant networks  $\net(x;\theta) =  \netTwo(\invarianceSet \cdot x;\theta)$ by simply using an 
$$\text{\textit{orbit mapping }} \orbitMapping:\{\invarianceSet \cdot x ~|~ x \in \inputSpace \} \rightarrow \inputSpace, $$
which uniquely selects a particular element from an orbit as a first layer in $\netTwo$. Subsequently, we can proceed with any standard network architecture and \cref{fact:invariance} still guarantees the desired invariance. A key in designing instances of orbit mappings is that they should not require enlisting all elements of $\invarianceSet \cdot x$ in order to evaluate $\orbitMapping(\invarianceSet \cdot x)$. Such constructions include, for instance, mappings that select elements from the orbit via an optimization problem such as 
\begin{align}
\label{eq:argmaxLayer}
  \orbitMapping(S\cdot x) = \argmax_{z \in S\cdot x} E(z),
\end{align}
for a suitable (application dependent) real-valued function $E$ (with some properties to ensure the existence and uniqueness of minimizers, and ideally a closed form solution). Note that the provable invariance holds even if $E$ itself is parameterized and thus learnable. Let us provide more concrete examples of orbit mappings. 

\begin{example}[Mean-subtraction]
A common approach in data classification tasks is to first normalize the input by subtracting its mean. Considering $\inputSpace = \R^n$ and $\invarianceSet = \{g:\R^n \rightarrow \R^n~|~ g(x) = x + a\mathds{1}, \text{ for some } a\in \R\} $, with $\mathds{1}\in \R^n$ being a vector of all ones, input-mean-subtraction is an orbit mapping that selects the unique element from any 
$\invarianceSet\cdot x$ which has zero mean. 
\end{example}
\begin{example}[Permutation invariance via sorting]
\label{example:sorting}
Consider $\inputSpace= \R^n$, and $\invarianceSet$ to be all permutations of vectors in $\R^n$, i.e., $\invarianceSet = \{s \in \{0,1\}^{n \times n} ~|~ \sum_i s_{i,j}=1~\forall j,~~ \sum_js_{i,j} = 1~\forall i\}$. We could define $E: \inputSpace \rightarrow \R$ via $E(x) = \sum_{i=1}^n \frac{1}{i} |x_i|, $
such that \eqref{eq:argmaxLayer} yields an orbit mapping that simply selects the element from an orbit whose entries are sorted by magnitude in an ascending order. 
\end{example}

With the very natural additional condition that our orbit mappings really select an element from the orbit, i.e., $\orbitMapping(\invarianceSet \cdot x) \in \invarianceSet \cdot x$, we can readily construct equivariant networks as well: 
\begin{proposition}[Orbit mapping for equivariant networks]
\label{fact:argmaxInvariance}
Let $\orbitMapping$ be an orbit mapping that satisfies  $\orbitMapping(\invarianceSet \cdot x) \in \invarianceSet \cdot x$ for all $x$. Any network $\net: \inputSpace \times \R^p \rightarrow \inputSpace$ that can be written as
\begin{align}
    \label{eq:equivariantArgmaxNets}
    \net(x;\theta) = \hat{g}^{-1}(\netThree(\hat{g}(x);\theta))
\end{align}
for an arbitrary network $\netThree: \inputSpace \times \R^p \rightarrow \inputSpace$ and $\hat{g} \in \invarianceSet$ denoting the element that satisfies $\hat{g}(x) = \orbitMapping(\invarianceSet \cdot x)$ is equivariant. 
\end{proposition}
We again refer to the appendix for a short proof.
Sticking to our \cref{example:sorting}, undoing the sort operation at the end of the network allows to transfer from an invariant, to an equivariant network. 

As a final note on the theory, our concept of orbit mappings can further be generalized by $\orbitMapping$ not mapping to the input space $\inputSpace$, but to a different representation, which can be beneficial for particular, complex groups. In geometry processing, for instance, an important group action are isometric deformations of shapes. A common strategy to handle these (c.f. \cite{ovsjanikov12functionalmaps}) is to identify any shape with the eigenfunctions of its Laplace-Beltrami operator  \cite{pinkall93cotan}, which represents a natural (generalized) orbit mapping. We refer to \cite{litany17deepFM, eisenberger2020deepshells, huang2019operatornet} for exemplary deep learning applications. 

\section{Applications}
We will now present two specific instances of orbit mappings for handling continuous rotations of images as well as for orientation and permutation invariant 3D point cloud classification.

\subsection{Invariance to continous image rotations}
\paragraph{Images as functions}
Let us consider the in practice important example of in- and equivariance to continuous rotations of images. To do so, consider $\inputSpace \subset \{u:\Omega\subset \R^2 \rightarrow \R\}$ to represent images as functions. For the sake of simplicity, we consider grayscale images only, but this extends to color images in a straight-forward way. In our notation $z \in \R^2$ represents spatial coordinates of an image (to avoid an overlap with our previous $x\in \inputSpace$, which we used for the input of a network). We set
\begin{align}
    \label{eq:rotationsOnImages}
    \begin{split}
    & \invarianceSet = \{g:\inputSpace \rightarrow \inputSpace ~|~ g(u)(z) = u\left(M(\alpha) z\right),
    \text{ for } \alpha \in \R \},\\
   & \text{and }  M(\alpha) = \begin{pmatrix}\sin(\alpha) & \cos(\alpha)\\ -\cos(\alpha) & \sin(\alpha) \end{pmatrix}.
    \end{split}
\end{align}
As $\invarianceSet$ has infinitely many elements, approaches that worked well for rotations by $90$ degrees like \cite{cohen2016group} are not applicable anymore. We therefore turn to \eqref{eq:argmaxLayer} and define 
\begin{align}
    E(u) = \left\langle \begin{pmatrix}1\\0 \end{pmatrix}, \int_Z \nabla u(z) ~dz  \right\rangle,
\end{align}
for a suitable set $Z$, e.g. a circle around the image center. In other words, we are proposing to look for the rotation $g\in \invarianceSet$ that makes the average gradient of the image over $Z$ point upwards. The attractive property of the above function $E$ is that rotations commute gradient operators such that
\begin{align}
\label{eq:gradRotation}
\begin{split}
    \hat{g} = \argmax_{g \in \invarianceSet}& \left\langle \begin{pmatrix}1\\0 \end{pmatrix}, \int_Z \nabla \hat{g}(u)(z) ~dz  \right\rangle \\
   \text{ is given in the form of \eqref{eq:rotationsOnImages} for }  \alpha &= 
    \arccos\left(\left\langle \begin{pmatrix}1\\0 \end{pmatrix}, \frac{\int_Z \nabla u(z) ~dz}{\|\int_Z \nabla u(z) ~dz\|}  \right\rangle\right).
\end{split}
\end{align}
Note that \eqref{eq:gradRotation} yields the unique solution to the maximization problem, unless $\int_Z \nabla u(z) ~dz=0$, in which case any $g \in \invarianceSet$ maximizes the objective, and a second criterion needs to select the unique orientation.  In practice, the magnitude of $\int_Z \nabla u(z) ~dz$ determines how stable the orbit mapping is, e.g. due to interpolation artifacts. 
 
\paragraph{Discretization}
\begin{figure}
    \centering
    \includegraphics[width=0.30\textwidth,angle=-90,origin=c,trim={0 0cm 0 0},clip]{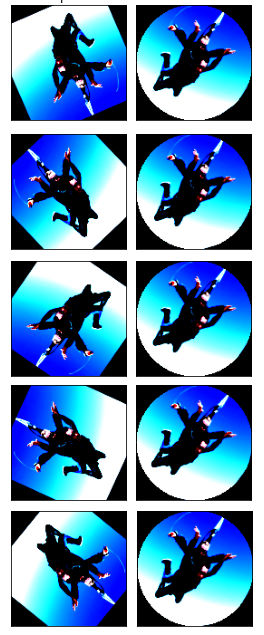}
    \vspace{-2.5cm}
    \caption{Images of different orientations (top) get aligned quite consistently with the proposed gradient-based strategy (bottom). }
    \label{fig:rotationIllustration}
\end{figure}
For a discrete (grayscale) image given a matrix $\tilde{u} \in \R^{n_y \times n_x}$, we first apply Gaussian blurring with a standard deviation of $\sigma = $ (to eliminate the effect of noise and create a smooth image), and subsequently construct an underlying continuous function $u:\Omega\subset \R^2 \rightarrow \R$ by bilinear interpolation. For the set $Z$ we choose two circles of radii $0.05$ and $0.4$ (for $\Omega$ being normalized to $[0,1]^2$). We approximate the integral by a finite sum over $1000$ evaluations of the derivative along each circle, using exact differentiation of the continuous image model. 
As illustrated in Fig.~\ref{fig:rotationIllustration}, this strategy can stabilize arbitrary rotations quite successfully. 
\begin{table*}
\centering
\resizebox{\linewidth}{!}{
\begin{tabular}{l l l l   l l l l l l}
    \hline
\multirow{2}{*}{Dataset}&\multirow{2}{*}{Eval.}&\multirow{2}{*}{Std.}&\multirow{2}{*}{RA} &\multirow{2}{*}{Advers.}&\multirow{2}{*}{Mixed}
&\multirow{2}{*}{Advers.-KL}&\multirow{2}{*}{Advers.-ALP} &Ours-Std & Ours-RA\\
&&&&&&&&(Train+Test) & (Train+Test)\\
\hline
\multirow{3}{*}{CIFAR10} & Clean.&\textbf{93.98}&85.54& 69.32&91.15
&72.28 &71.25 &87.99&85.40\\
& Avg.&40.06&75.99&68.54&68.37
&70.29&70.30&\textbf{84.12}&81.82\\
&Worst &1.31&44.71&50.21&17.15
&51.05&52.29&68.60&\textbf{71.09}\\
\hline
\multirow{3}{*}{HAM10000}&Clean&\textbf{93.82}&93.30&92.28
&93.71&92.54&92.92&93.31&93.41\\
&Avg.&91.73&90.81&91.87&\textbf{92.13}&91.79
&91.85&91.38&\textbf{92.13}\\
&Worst&82.52&82.30
&85.04&84.53&85.92&86.38&87.96&\textbf{88.55}\\
\hline
\multirow{3}{*}{CUB200}&Clean&\textbf{77.41}&69.89&64.54&68.56
&64.47&64.63&71.19&69.55\\
&Avg.&52.45 &70.12&64.07&65.91
&64.65&64.34&\textbf{71.56}&69.59\\
&Worst.&8.07&41.01&42.82&42.87
&43.04&43.63&\textbf{58.80}&56.88\\
\hline
\end{tabular}}
\vspace{0.5pt}
\caption{Comparing rotational invariance using training schemes vs. analytical element selection.Shown are the mean clean accuracy and the average and worst case accuracies when test images are rotated in steps of 1 degree using bilinear interpolation. The mean values over 5 runs are reported are reported.}
\label{tab:rob.train_reg.}
\end{table*}

\begin{table*}
\centering
\resizebox{\linewidth}{!}{
\begin{tabular}{l l  l  l l l l l  c l l l }
    \hline
\multirow{2}{*}{Augment} &  \multirow{2}{*}{OM}& \multirow{2}{*}{Clean}&  \multicolumn{3}{c}{Average }& \multicolumn{3}{c}{Worst-case}\\
\cline{4-6}\cline{8-10}
&&&Nearest&Bilinear&Bicubic&&Nearest&Bilinear&Bicubic\\
\hline
\multirow{2}{*}{Std.}  &  \ding{55} & \textbf{93.98$\pm$0.32} & 35.12$\pm$0.81& 40.06$\pm$0.44 &42.81$\pm$0.50&&0.79$\pm$0.38 &1.31$\pm$0.13& 2.22$\pm$0.17\\
&\ding{51} Train+Test& 87.99$\pm$0.43&72.40$\pm$0.33& 84.12$\pm$0.55 &\textbf{86.61$\pm$0.49}&&34.57$\pm$0.94 &68.60$\pm$0.81& 74.49$\pm$0.84\\

\hline 
\multirow{3}{*}{RA}& \ding{55} &
85.54$\pm$0.72&80.47$\pm$0.74& 75.99$\pm$0.72 &79.47$\pm$0.65&&45.50$\pm$0.83 &44.71$\pm$0.74& 50.50$\pm$0.78\\
 & \ding{51} Test    &79.26$\pm$0.42
 &	74.93$\pm$0.51& 69.31$\pm$0.65 &73.94$\pm$0.63&&48.93$\pm$0.75 &52.18$\pm$0.91& 58.69$\pm$0.78\\
& \ding{51} Train+Test& 85.40$\pm$0.57&\textbf{84.37$\pm$0.58}& 81.82$\pm$0.59 &84.82$\pm$0.52&&\textbf{66.22$\pm$0.75} &71.09$\pm$1.01& 76.44$\pm$0.89\\
\hline
\multirow{3}{*}{\shortstack{RA-\\combined}}      &\ding{55}&92.42$\pm$0.21&80.90$\pm$0.64& 82.23$\pm$0.74 &82.71$\pm$0.69&&36.98$\pm$1.27 &48.07$\pm$1.66& 49.51$\pm$1.47\\
&  \ding{51} Test&   82.55$\pm$0.86&76.33$\pm$0.95& 77.93$\pm$0.68 &78.42$\pm$0.64&&45.44$\pm$1.32 &60.23 $\pm$1.24&62.18$\pm$1.33\\
&  \ding{51} Train+Test&86.69$\pm$0.12&84.06$\pm$0.21& \textbf{85.27$\pm$0.23} &86.06$\pm$0.20&&61.75$\pm$0.76 &\textbf{75.29$\pm$0.42}& \textbf{77.25$\pm$0.27}\\
\hline     
Advers.& \ding{55}&69.32$\pm$1.61&61.73$\pm$1.12& 68.54$\pm$0.68 &68.00$\pm$0.31&&36.95$\pm$0.97 &50.21$\pm$0.55& 49.73$\pm$0.98\\
Mixed& \ding{55}&91.15$\pm$0.15&54.55$\pm$0.40& 68.37$\pm$0.66 &68.48$\pm$0.37&&3.86$\pm$0.13 &17.15$\pm$1.25& 16.85$\pm$0.93\\
Advers.-KL& \ding{55}&72.28$\pm$2.05&62.60$\pm$1.72& 70.29$\pm$1.42 &69.84$\pm$1.29&&32.60$\pm$0.74 &51.05$\pm$2.47& 51.11$\pm$1.03\\
Advers.-ALP&\ding{55}&71.25$\pm$0.97&62.36$\pm$2.19& 70.30$\pm$1.50 &69.71$\pm$1.22&&33.98$\pm$1.44 &52.29$\pm$1.76& 52.57$\pm$1.57\\
\hline
\end{tabular}
}
\vspace{0.5pt}
\caption{Effect of augmentation and including gradient based orbit mapping \textit{(OM)} on robustness to rotations with different interpolations for CIFAR10 classification using Resnet-18. Shown are clean accuracy on standard test set and  average and worst-case accuracies on rotated test set. Mean and standard deviations over 5 runs are reported. }
\label{tab:train_test}
\end{table*}

\paragraph{Experiments} Recent works show that adversarial training~\cite{engstrom2019exploring} and regularization~\cite{yang2019invariance} improve robustness to spatial transformations. These works restrict themselves to smaller transformation sets, as images have a natural orientation, and invariance to  the whole orbit is not needed.  However, exact invariance to continuous rotations is desirable, e.g. in certain medical image classification tasks \cite{finlayson2019adversarial}. 
To evaluate our approach, we apply our rotational alignment module to image classification networks on three different datasets: On CIFAR-10 we train a Resnet-18 \cite{he2016deep} from scratch. On the HAM10000 skin image dataset \cite{tschandl2018ham10000} we finetune an NFNet-F0 network~\cite{brock2021high}, and on CUB-200~\cite{wah2011caltech} we finetune a Resnet-50~\cite{he2016deep}, both of which were pretrained on ImageNet. While the datasets CIFAR10 and CUB-200 have an inherent variance in orientation, for the skin lesion classification on HAM10000 exact rotation invariance is desirable. The details of the protocol used for training all our networks as well as some additional experiments are provided in the supplementary material. We compare with following approaches:\\
\emph{i)~adversarial training:}
$\min_\theta \sum_{\text{examples i}} \loss(\net(\hat{x}^i;\theta);y^i),~\text{for}~\hat{x}^i=\argmax_{z \in S\cdot x^i}\loss(\net(z);y^i)$\\
This is approximated by selecting the worst out of  10 different random rotations for each image  in  every iteration, as proposed in \cite{engstrom2019exploring}. It is referred to as Advers. in  Tab.~\ref{tab:rob.train_reg.}.\\
\emph{ii)~mixed mode training:}  $\min_\theta \sum_{\text{examples i}} \loss(\net(\hat{x}^i;\theta);y^i)+\loss(\net(x^i;\theta);y^i)$ which uses both natural and adversarial examples $\hat{x}^i$. In addition, we compare with the use of following regularizers~\cite{yang2019invariance} in conjunction with adversarial training:\\
$\text{ } $ ~ \emph{a)~adversarial logit pairing~(ALP)}:  $R_{ALP}(\net,x^i,y^i)=\|\net(x^i;\theta)-\net(\hat{x}^i;\theta)\|_2^2$ , \\
$\text{ } $ ~ \emph{b)~KL-divergence}:$R_{KL}(\net,x^i,y^i)=D_{KL}(\net(x^i;\theta)||\net(\hat{x}^i;\theta))$ . \\
The use of these regularizers in addition to adversarial training is indicated as Advers.-ALP and Advers.-KL in Tab.~\ref{tab:rob.train_reg.}. We also compare with the simple baseline of augmenting all the images with random rotations, referred to as RA in Tab.~\ref{tab:rob.train_reg.}. We would like to point out that adversarial training using the worst of $10$ samples roughly increases the training effort of the underlying model by a factor of~$5$. 

\paragraph{Results}

We measure the accuracy each model achieves on the original dataset without rotations (\textit{clean}), as well as the average (\textit{Avg.}) and  worst-case (\textit{Worst}) accuracies in the range of whole orbit of rotations discretized in steps of of 1 degree, where `\textit{Worst}' counts an image as misclassified as soon as there exists a rotation at which the network makes a wrong prediction. 

As we can see in Tab.~\ref{tab:rob.train_reg.}, networks trained without rotation augmentation perform poorly in terms of both, the average and worst-case accuracy, if the data set contains an inherent orientation. While augmenting with rotations during training results in improvements, there is still a huge gap ($\sim 30\%$ for CIFAR10 and CUB200) between the average and worst-case accuracies. {While adversarial training approaches improve the performance in the worst case, there is a clear drop in the clean and average accuracies when compared to data augmentation.} In contrast, our approach of rotating the images to align with the gradient significantly improves the worst case accuracy even without augmenting with rotations, with performance  better than the adversarial training~\cite{engstrom2019exploring} and regularization approaches~\cite{yang2019invariance}. When the data itself does not contain a prominent orientation as in the HAM10000 data set, the general trend in accuracies still holds (\emph{clean$>$average$>$worst-case}), but the drops in accuracies are not drastic, and adversarial training schemes/ orbit mapping provide comparable improvements in robustness. 

\emph{Discretization Artifacts}\hspace{.3cm} It is interesting to see that while consistently selecting a single element from the continuous orbit of rotations leads to provable rotational invariance when considering images as continuous functions, discretization artifacts and boundary effects still play a crucial in practice, and rotations cannot be fully stabilized. As a result there is still discrepancy between the average and worst case accuracies, and the performance is further improved when our approach also uses rotation augmentation. Motivated by the strong effect the discretization seems to have, we investigate different interpolation schemes used to rotate the image in more detail: Tab.~\ref{tab:train_test} shows the results different training schemes with and without our orbit mapping (\textit{OM}) obtained with a ResNet-18 architecture on CIFAR-10 when using different types of interpolation. Besides standard training without augmentation (\textit{Std.}), we use rotation augmentation (\textit{RA}) using the Pytorch-default of nearest-neighbor interpolation, a combined augmentation scheme (\textit{RA-combined}) that applies random rotation only to a fraction of images in a batch using at least one nearest neighbor, one bilinear and one bicubic interpolation, and the adversarial training and regularization from \cite{engstrom2019exploring,yang2019invariance} using bilinear interpolation (following the authors' implementation). 

Most notably, the worst-case accuracies between different types of interpolation may differ by more than $20\%$, indicating a huge influence of the interpolation scheme. Adversarial training with bi-linear interpolation still leaves a large vulnerability to image rotations with nearest neighbor interpolation. While our approach without rotation augmentation is also vulnerable to interpolation effects, it is ameliorated when using rotation augmentation. We observe that including different augmentations (RA-combined) improves the robustness significantly. Additionally applying our orbit mapping at test time further improves the results and clearly outperforms the robust training and regularization approaches for CIFAR-10.

\subsection{Permutation, Rotation, and Scaling Invariance in 3D Point Cloud Classification}

Another example of a desired invariance is the ability of networks to classify objects given as $3$D point clouds independent of their orientation and scale. 
Popular modern architectures, such as PointNet \cite{qi2017pointnet} and subsequent improvements \cite{qi2017pointnetpp}, to rely on the ability of modern neural networks to learn such invariances from data, through training on large datasets and extensive data augmentations. 
PointNet includes two learnable spatial-transformer components that have the capacity to learn these invariances. 
However, this results in a significant increase of variables in the architecture, where two-thirds of all learnable parameters are contained in the spatial transformer components. 
Furthermore, it is not clear a-priori how much training is required for the network to be, even on average, robust against these transformations. 
We discuss this question, and supplement it with experiments showing that PointNet performs better with built-in invariance via our orbit mappings than with augmentation only on the \textit{modelnet40} dataset \cite{wu2015shapenets}.

In this setting, $\inputSpace = \R^{d \times N}$ are $N$ many $d$-dimensional coordinates (usually with $d=3$), and the desired group actions to be invariant against are (left-)multiplication with a 
rotation matrix, as well as multiplication with any number $c\in\R^+$ to account for different scaling. 

\paragraph{Scaling}
Desired invariances in point cloud classification range from class-dependent variances to geometric properties. 
For example, the classification of airplanes should be invariant to the specific wing shape, as well as the scale or translation of the model. 
Networks can obtain invariance against both with enough training data; however, our experiments show that even simple transformations like scaling and translation are not learned robustly outside the scope of what was provided in the training data, see Tab.~\ref{tab:3d_invariances}.
This is surprising when considering that both can be undone by centering around the origin and re-scaling. 
Invariance to scaling can be achieved in the sense of Sec.~\ref{sec:proposedApproach} by scaling input point-clouds by the average distance of all points to the origin. Our experiments show that this leads to robustness against much more extreme transformation values without the need for expensive training, both for average as well as worst-case accuracy. 
We tested the worst-case accuracy on the following scales: $\lbrace 0.001, 0.01, 0.1, 0.5, 1.0, 5.0, 10, 100, 1000 \rbrace$. 
While our approach performs well on all cases, training PointNet on random data augmentation in the range of possible values actually reduces the accuracy on clean, not scaled test data. 
If the training is only done on a subset of the interval, the average and worst-case robustness decreases significantly.
See Tab.~\ref{tab:3d_invariances}.

\paragraph{Rotation}
Here, we show that 3D rotations lead to a similar behavior and should also be treated via an architecture that implies the desired invariance. 
In our framework, we choose the element of the orbit to be the rotation of $\inputSpace$ that aligns its principle components with the coordinate axes. 
The optimal transformation can be found via the singular value decomposition $X=U\Sigma V$ of the point cloud $X$ up to the arbitrary orientation of the principle axes. 
To remove the ambiguity, we choose signs of the first row of $U$ and encode them into a diagonal matrix $D$, such that the final transform is given by $\hat{X} = XV^\top D$.

We apply this rotational alignment to PointNet and evaluate its robustness to rotations in average-case and worst-case when rotating the validation dataset in $16 \times 16$ increments (i.e. with $16$ discrete angles along each of the two angular degrees of freedom of a 3D rotation). 
Tab.~\ref{tab:3d_invariances} shows that PointNet trained without augmentation is susceptible in worst-case and even average-case rotations. 
This can be ameliorated in the average-case by training with random rotations, but the worst-case accuracy is still significantly low. 
On the other hand, explicitly stabilizing rotations at test time by PCA does provide effortless invariance to rotations. 

An interesting subtle observation is that the PCA does not entirely replace data augmentations, as the best accuracy is reached when using data augmentations during training and PCA during testing - the model is improved through more data via seeing random rotations of objects during training, but then made robust by testing on aligned object point clouds.

\begin{table}
\small
\centering
    \resizebox{\linewidth}{!}{\begin{tabular} {c c}
\begin{tabular}{ c c | l l l}

\multicolumn{5}{c}{\textbf{Scaling Invariance:}}\\
\hline
Augmentation & Unscaling & Clean & Avg. & Worst \\
 $[0.8, 1.25]$ & \ding{55} & \textbf{85.96} & 24.05 & 00.00 \\ 
 $[0.01, 100]$ & \ding{55} & 82.35 & 48.60 & 00.00 \\ 
 $[0.001, 1000]$ & \ding{55} & 30.69 & 41.14 & \phantom{0}2.33\\
  $[0.001, 1000]$  & \ding{51}(Train+Test) & 85.77 & 85.77 & 85.77\\
 - & \ding{51}(Train+Test) & 86.10 & \textbf{86.10} & \textbf{86.10 }\\ 
 $[0.8, 1.25]$ & \ding{51}(Test) &  85.77 & 85.47 & 85.47 \\

\hline
\end{tabular} 
&       
\begin{tabular}{ c c | l l l}
\multicolumn{5}{c}{\textbf{Rotational Invariance:}}\\
\hline
 Augmentation & PCA & Clean & Avg. & Worst \\
 \ding{55} & \ding{55} & \textbf{85.96} &10.58 & \phantom{0}0.15 \\
 \ding{51} & \ding{55} & 71.87 & 72.85 & 28.86 \\
  \ding{51} & \ding{51}(Train+Test) & 72.25 & 72.25 & 72.25 \\
 \ding{55} & \ding{51}(Train+Test) & 73.51 & 73.51 & 73.51 \\
 \ding{51} & \ding{51}(Test) & 78.45 & \textbf{78.45} & \textbf{78.45} \\
 & & & & \\
 \hline
\end{tabular}
\end{tabular}}
\vspace{0.5pt}
\caption{Invariances in 3D pointcloud classification. Experiments with PointNet trained for classification of modelnet40 data, with and without data augmentation or analytical inclusion of each invariance. Shown is class accuracy of the final iterate on modelnet40 for the validation dataset \textit{(Clean)}, and  class accuracy in average- \textit{(Avg.)} and worst-case \textit{(Worst)} for scaled (left) and rotated (right) evaluations of each pointcloud in the validation set.\label{tab:3d_invariances}}
\end{table}

\section{Limitations}
While we have shown the efficacy of our approach for the average and worst-case accuracy on both images and $3D$ point clouds, the average and worst-case values do not coincide for images, and a large discrepancy between the theoretical provable invariance (in the perspective of images as continuous functions) and the practical discrete setting remains. We conjecture that this is related to discretization artifacts when applying rotations that change the gradient directions, especially at low resolutions.
Notably, such artifacts appear much more frequently in artificial settings, e.g. during data augmentation or when testing for worst-case accuracy, than in photographs of rotating objects that only get discretized once. 
\section{Conclusion} 
We propose a simple and general way of incorporating invariances to group actions in  neural networks, by uniquely selecting a specific element from the orbit of group transformations. This guarantees provable invariance to group transformations for 3D point clouds, and  demonstrates significant improvements in worst case accuracies with respect to continuous rotations of images over adversarial training with a limited computational overhead. To draw conclusions about the question raised in our title, we found a consistent advantage of enforcing the desired invariance via the architecture rather than the training, with additional advantages arising from their combination by exposing the model to a large variety of examples: Training with data augmentation and orbit mappings (in cases where discretization artifacts prevent a provable invariance of the latter), or utilizing data augmentation during training and enabling orbit mappings during inference, results in the most accurate predictions. 
{\small
\bibliographystyle{ieee_fullname}

}
\appendices
\section{Proofs of Propositions 1 and 2}
\begin{proof}[Proof of Proposition 1]
We want to show that a network satisfying the 
condition (5) is equivariant. Let $h \in S$ be arbitrary. Note that  
\begin{align}
    \{g ~|~ g\in  \invarianceSet\} = 
    \{h^{-1}g ~|~ g \in \invarianceSet\} 
\end{align}
such that a substitution of variables from $g \in \invarianceSet$ to $z = h^{-1}g \in \invarianceSet$ (i.e., $g = hz$ and $z^{-1} = g^{-1}h$) yields
\begin{align*}
    &\{g (\netThree (g^{-1}(h(x));\theta_2)) ~|~ g \in \invarianceSet\} \\
    =& 
    \{h( z (\netThree (z^{-1}(x);\theta_2))) ~|~ z \in \invarianceSet\}.
\end{align*}
This means that we can also write  \begin{align*}
    \net(h(x);\theta) &=  \netTwo(\{h( z (\netThree (z^{-1}(x);\theta_2))) ~|~ z \in \invarianceSet\};\theta_1)\\
    &=  \netTwo(h(\{z (\netThree (z^{-1}(x);\theta_2)) ~|~ z \in \invarianceSet\});\theta_1)\\
    &=  h(\netTwo(\{z (\netThree (z^{-1}(x);\theta_2)) ~|~ z \in \invarianceSet\});\theta_1)\\
    &= h(\net(x;\theta))
\end{align*}
which yields the desired equivariance under the assumed commutative property.
\end{proof}

\begin{proof}[Proof of Proposition  2]
We want to show that a network satisfying the 
condition (7) is equivariant. Consider an input $a = r(x)$ to the network, where $r$ denotes an arbitrary element of $S$. We first need to determine the element $\tilde{g} \in S$ such that $\tilde{g}(a) = h(S\cdot a)$. From the definition of the orbit, it follows that $S \cdot x = S \cdot r(x)$, such that our orbit mapping satisfies remains the same, i.e., $h(S \cdot x) = h(S \cdot a) = \hat{g}(x)$. Solving the equation  $\tilde{g}(a) = \hat{g}(x)$ with $a = r(x)$, i.e., $x = r^{-1}(a)$ for $\tilde{g}$ yields $\tilde{g} = \hat{g} r^{-1}$. Now it follows that
\begin{align*}
\net(r(x);\theta) = \net(a;\theta) &= \tilde{g}^{-1}(\netThree( \tilde{g}(a);\theta))\\
& = r(\hat{g}^{-1}(\netThree( \tilde{g}(a);\theta)))\\
& = r(\hat{g}^{-1}(\netThree(\hat{g}(x);\theta)))\\
& = r(\net(x;\theta)),
\end{align*}
which concludes the proof.
\end{proof}

\section{A Discussion on Isometry Invariance}

Here, we will elaborate on how the functional map framework \cite{ovsjanikov12functionalmaps} can be seen as an application of our orbit mapping for isometry invariance.
Functional maps are a widely used method to find correspondences between isometric shapes, and we will show here that the framework fits within our proposed theory. 
Non-rigid correspondence is a notoriously hard problem, and joint optimization within a larger framework makes it even more complex.
To resolve this the idea of functional maps is to change the representation of the correspondence from point-wise to function-wise. 
By choosing the eigenfunctions of the Laplace-Beltrami operator \cite{pinkall93cotan}  as the basis for functions on the shapes, the problem becomes a least squares problem aligning suitable descriptor functions in the space of functions. 

Here, $F \in \mathcal{F}(\mathcal{X})$ and $G \in \mathcal{F}(\mathcal{Y})$ are descriptor functions on the shapes $\mathcal{X}$ and $\mathcal{Y}$ respectively. They are assumed to take similar values on corresponding points on $\mathcal{X}, \mathcal{Y}$, and generate the designated orbit element within our framework. 
These descriptors are projected onto the eigenfunctions of $\mathcal{X}, \mathcal{Y}$, named $\Phi, \Psi$ respectively. 
These projections are the chosen elements of the orbit we will align, and, for isometries and sufficiently comparable descriptors, the projections can be aligned by an orthogonal transformation generating the group action which is exactly the functional map $C$.
The vanilla functional map optimization looks like this:

\begin{align}
    \underset{C \in O(k)}{\arg\min} \Vert C \Phi^{-1} F - \Psi^{-1} G \Vert_2^2
\end{align}

Functional maps are often used when shape correspondence is required within another framework, and has been used in many deep learning applications \cite{litany17deepFM, eisenberger2020deepshells, huang2019operatornet}. 
Due to its wide application, we will not provide extra experiments to show its efficacy but want to emphasize that this is a possible implementation of our theory. 

\section{Details about the Experimental Setting}
In the following we provide the detailed training settings used in our experiments.
\subsection{Rotation invariance for images}
For our experiments with image rotational invariance, we used  Pytorch(v.1.8.1), python(v.3.8.8), torchvision(v.0.9.1). The exact training protocol is provided below.
\paragraph{CIFAR10} We trained a Resnet18~\cite{he2016deep} on the CIFAR 10 dataset, using stochastic gradient descent with initial learning rate 0.1, momentum 0.9, and weight decay 5e-4. Additionally, we trained a small Convnet and a linear model which used an initial learning rate of 0.01.  For all the models, the learning rate is decayed by a factor of 0.5 whenever the validation loss does not decrease for 5 epochs.  Training data is augmented using random horizontal flips, random crops of size 32 after zero-padding by 4 pixels. We divide the training data into train (80\%) and validation (20\%) sets. Networks are trained for 150 epochs with a batch size of 128 and we report the results on the test set using the model with best validation accuracy. The experiments with CIFAR10 were performed partially on a machine with one Nvidia TITAN RTX, and partially on machine with 4 NVIDIA GeForce RTX 2080  GPUs.
\paragraph{HAM10000} We fine-tuned an imagenet pretrained\footnote{ pretrained model from \url{https://github.com/rwightman/pytorch-image-models} licensed Apache 2.0} NFNet-F0~\cite{brock2021high} on HAM10000 dataset \cite{tschandl2018ham10000}. The dataset is split into  8912 train and 1103 validation images using stratified split, ensuring there are no duplicates with the same lesion ids in the train and validation sets. Training data is augmented using random horizontal and vertical flips and color jitter, and randomly oversample the minority classes to mitigate class imbalance.  The network is finetuned for 10 epochs, with a batch size of 128 and learning rate  of 1e-4 using Adam optimizer~\cite{kingma2014adam}  with exponential learning rate decay, with factor 0.95. We report results using final iterate on the validation set. The experiments with HAM10000 dataset were partially performed on a machine with one NVIDIA TITAN RTX card, and partially on machine with 4 NVIDIA GeForce RTX 2080  GPUs.
\paragraph{CUB200}
This is a small  dataset  containing 11,788 images of birds, split into 5994 images for training and 5794 test images. Since training a network from scratch gives low accuracies (around 35\% clean accuracy with Resnet-50), we instead perform finetuning using an imagenet pretrained Resnet-50 from pytorch torchvision~(v.0.9.1)  on CUB-200 dataset ~\cite{wah2011caltech}. The training data is augmented using random horizontal flips, random resized crops of size 224.  The network is finetuned for 60 epochs with batch size of 128 and initial learning rate of 1e-4, using Adam optimizer~\cite{kingma2014adam} with exponential learning rate decay, with factor 0.95. We report the accuracies using the final iterate on the test set. The experiments on CUB-200 dataset were performed on machine with 4 NVIDIA GeForce RTX 2080  GPUs. 

All the  three image  datasets including  HAM10000 dataset \cite{tschandl2018ham10000}  used in our experiments are publicly available and widely used in machine learning literature. To the best of our knowledge these do not contain offensive content or personally identifiable information. 

\subsection{Rotation and Scale invariance for 3D point clouds}
We investigate invariance to rotations and scale for 3D point clouds with the task of point cloud classification on the \textit{modelnet40} dataset \cite{wu20153d}. For this dataset note the asset descriptions at \url{https://modelnet.cs.princeton.edu/}: "All CAD models are downloaded from the Internet and the original authors hold the copyright of the CAD models. The label of the data was obtained by us via Amazon Mechanical Turk service and it is provided freely. This dataset is provided for the convenience of academic research only." We use the resampled version of \url{shapenet.cs.stanford.edu/media/modelnet40_normal_resampled.zip}. We follow the hyperparameters of \cite{qi2017pointnet,qi2017pointnetpp} with improvements from the implementation of \cite{yan_pointnet} on which we base our experiments. We train a standard PointNet for 200 epochs with a batch size of 24 with Adam \cite{kingma2014adam} with base learning rate of $0.001$, weight decay of $0.0001$. During training we sample 1024 3D points from every example in \textit{modelnet40}, randomly scale with a scale from the interval $[0.8, 1.25]$, and randomly translate by an offset of up to $0.1$ - if not otherwise mentioned in our experiments. This is the training procedure proposed in \cite{yan_pointnet}. However, we always train the model for the the full 200 epochs and report final \textit{class} accuracy based on the final result - we do not report instance accuracy. We further report invariance tests based on the final model. 

As described in the main body, we evaluate rotational invariance by testing on $16\times 16$ regularly spaced angles from $[0, 2\pi]$, rotating along $xy$ and $yz$ axes. We evaluate scaling invariance by testing the scales $\lbrace 0.001, 0.01, 0.1, 0.5, 1.0, 5.0, 10, 100, 1000\rbrace$.
All experiments for this dataset were run on three single GPU office machines, containing an NVIDIA TITAN Xp, and two GTX 2080ti, respectively.
\section{Additional Numerical Results}
\begin{table*}
\centering
\resizebox{\linewidth}{!}{
\begin{tabular}{l l l l l l  c l l l }
    \hline
 Train& OM &Clean.&  \multicolumn{3}{c}{Average}& \multicolumn{3}{c}{Worst-case}\\
\cline{4-6}\cline{8-10}
&&&Nearest&Bilinear&Bicubic&&Nearest&Bilinear&Bicubic\\
\hline
 \multirow{2}{*}{Std.}  &\ding{55}&\textbf{77.41$\pm$0.33}&37.67$\pm$0.35& 52.45$\pm$0.29 &51.87$\pm$0.31&&3.19$\pm$0.49 &8.07$\pm$0.35& 8.16$\pm$0.33
\\
&\ding{51}Train+Test&71.19$\pm$0.34&63.35$\pm$0.30& \textbf{71.56$\pm$0.34} &\textbf{70.93$\pm$0.35}&&40.63$\pm$0.48 &\textbf{58.80$\pm$0.39}&\textbf{ 59.02$\pm$0.41}\\
\hline
 \multirow{3}{*}{RA.}  &\ding{55}&69.89$\pm$0.28&67.61$\pm$0.33& 70.12$\pm$0.34 &68.83$\pm$0.37&&34.88$\pm$0.47 &41.01$\pm$0.41& 40.50$\pm$0.43
\\
&\ding{51}Test&69.41$\pm$0.31&\textbf{69.19$\pm$0.32}& 69.27$\pm$0.29 &68.53$\pm$0.38&&\textbf{48.63$\pm$0.43} &56.28$\pm$0.39& 55.86$\pm$0.40\\
&\ding{51}Train+Test&69.55$\pm$0.31&68.27$\pm$0.34& 69.59$\pm$0.22 &68.82$\pm$0.28&&48.54$\pm$0.39 &56.88$\pm$0.26& 56.77$\pm$0.31 \\
\hline
Advers.&\ding{55}& 64.54$\pm$0.17&53.74$\pm$0.65& 64.07$\pm$0.25 &63.22$\pm$0.54&&26.63$\pm$0.79 &42.82$\pm$0.60& 42.44$\pm$0.55
\\
 Mixed &\ding{55}&
68.56$\pm$0.46&57.17$\pm$0.60& 65.91$\pm$0.42 &65.76$\pm$0.51&&28.06$\pm$0.58 &42.87$\pm$0.32& 42.92$\pm$0.38
\\
Advers.-KL&\ding{55}&
64.47$\pm$0.35&53.93$\pm$0.35& 64.65$\pm$0.26 &64.02$\pm$0.34&&26.94$\pm$0.46 &43.04$\pm$0.63& 42.61$\pm$0.37
\\
Advers.-ALP&\ding{55}&
64.63$\pm$0.31&55.56$\pm$0.67& 64.34$\pm$0.17 &63.21$\pm$0.24&&29.55$\pm$0.69 &43.63$\pm$0.21& 43.48$\pm$0.32
\\
\hline
\end{tabular}}
\vspace{0.5pt}
\caption{Effect of augmentation and including gradient based orbit mapping \textit{(OM)} on robustness to rotations with different interpolations for CUB200 classification using Resnet50. Shown are clean accuracy on standard test set and  average and worst-case accuracies on rotated test set. Mean and standard deviations over 5 runs are reported. }
\label{tab:rob.train_reg_cub.}
\end{table*}

\begin{table*}
\centering
\resizebox{\linewidth}{!}{
\begin{tabular}{l |l l l    l l l  c l l l }
    \hline

Network &  Train& OM& Std.&  \multicolumn{3}{c}{Average}& \multicolumn{3}{c}{Worst-case}\\
\cline{5-7}\cline{9-11}
&&&&Nearest&Bilinear&Bicubic&&Nearest&Bilinear&Bicubic\\
\hline
\multirow{6}{*}{Linear}    & \multirow{2}{*}{Std.}      &\ding{55}&  \textbf{38.89$\pm$0.17}&25.31$\pm$0.21& 25.57$\pm$0.22 &25.48$\pm$0.24&&2.50$\pm$0.11 &3.56$\pm$0.17& 3.26$\pm$0.11
\\
  &     &\ding{51}Train+Test& 31.87$\pm$0.10&\textbf{31.25$\pm$0.04}& \textbf{31.58$\pm$0.05} &\textbf{31.33$\pm$0.04}&&13.08$\pm$0.23 &18.85$\pm$0.21& 18.21$\pm$0.21 \\

\cline{2-11}

&\multirow{3}{*}{RA}  &\ding{55}& 29.73$\pm$0.18&30.66$\pm$0.03& 30.77$\pm$0.03 &30.72$\pm$0.03&&14.30$\pm$0.42 &18.31$\pm$0.29& 16.94$\pm$0.37\\ 
&    & \ding{51}Test  &	30.60$\pm$0.13&30.52$\pm$0.07& 30.65$\pm$0.08 &30.54$\pm$0.09&&16.83$\pm$0.47 &21.17$\pm$0.28& 20.37$\pm$0.26\\
&    & \ding{51}Train+Test  &31.06$\pm$0.26&31.07$\pm$0.11& 31.27$\pm$0.10 &31.13$\pm$0.09&&\textbf{19.19$\pm$0.28} &\textbf{24.25$\pm$0.31}&\textbf{23.68$\pm$0.31}\\
\cline{2-11}
&  Advers.         &\ding{55}&28.82$\pm$0.77&29.46$\pm$0.60& 29.62$\pm$0.56 &29.36$\pm$0.56&&11.45$\pm$0.81 &14.20$\pm$0.93& 13.65$\pm$0.55\\ 
\hline
\hline

\multirow{6}{*}{Convnet} &\multirow{2}{*}{Std.}     &\ding{55} &
\textbf{86.12$\pm$0.33}&32.01$\pm$0.32& 35.97$\pm$0.26 &38.15$\pm$0.36&&0.85$\pm$0.09 &0.57$\pm$0.06& 0.89$\pm$0.14
\\
&&\ding{51}Train+Test &76.13$\pm$0.96&64.34$\pm$0.35& 71.21$\pm$0.96 &\textbf{74.61$\pm$0.84}&&25.78$\pm$0.49 &49.60$\pm$0.79& 55.57$\pm$0.81\\

                          

\cline{2-11}   

                          & \multirow{3}{*}{RA}    &\ding{55} &75.03$\pm$0.99&71.77$\pm$0.84& 65.45$\pm$0.66 &70.22$\pm$0.66&&27.96$\pm$0.50 &27.06$\pm$0.61& 32.51$\pm$0.53\\ 
                        &    & \ding{51}Test &70.12$\pm$0.64&67.64$\pm$0.55& 61.03$\pm$0.67 &66.09$\pm$0.71&&39.01$\pm$0.57 &42.88$\pm$0.90& 49.39$\pm$0.68\\
                          &    & \ding{51}Train+Test &74.30$\pm$0.77&\textbf{73.24$\pm$0.58}& 69.52$\pm$0.53 &73.38$\pm$0.59&&\textbf{46.25$\pm$0.54} &\textbf{53.36$\pm$0.57}& \textbf{59.04$\pm$0.53}\\

\cline{2-11}   
& Advers.         &\ding{55}  &72.96$\pm$0.95&62.08$\pm$0.59& \textbf{74.29$\pm$0.88} &73.86$\pm$0.76&&26.24$\pm$0.43 &50.99$\pm$0.54& 52.46$\pm$0.51\\ 


\hline
\hline
\multirow{6}{*}{Resnet18} &
\multirow{2}{*}{Std.}  &  \ding{55} & \textbf{93.98$\pm$0.32} & 35.12$\pm$0.81& 40.06$\pm$0.44 &42.81$\pm$0.50&&0.79$\pm$0.38 &1.31$\pm$0.13& 2.22$\pm$0.17\\
&&\ding{51} Train+Test& 87.99$\pm$0.43&72.40$\pm$0.33& \textbf{84.12$\pm$0.55} &\textbf{86.61$\pm$0.49}&&34.57$\pm$0.94 &68.60$\pm$0.81& 74.49$\pm$0.84\\

\cline{2-11} 
&\multirow{3}{*}{RA}& \ding{55} &
85.54$\pm$0.72&80.47$\pm$0.74& 75.99$\pm$0.72 &79.47$\pm$0.65&&45.50$\pm$0.83 &44.71$\pm$0.74& 50.50$\pm$0.78\\
& & \ding{51} Test    &79.26$\pm$0.42
 &	74.93$\pm$0.51& 69.31$\pm$0.65 &73.94$\pm$0.63&&48.93$\pm$0.75 &52.18$\pm$0.91& 58.69$\pm$0.78
\\
&&  \ding{51} Train+Test& 85.40$\pm$0.57&\textbf{84.37$\pm$0.58}& 81.82$\pm$0.59 &84.82$\pm$0.52&&\textbf{66.22$\pm$0.75} &\textbf{71.09$\pm$1.01}&\textbf{76.44$\pm$0.89}\\
\cline{2-11} 
\comment{&\multirow{3}{*}{\shortstack{RA$-$\\combined}}      &\ding{55}&92.42$\pm$0.21&80.90$\pm$0.64& 82.23$\pm$0.74 &82.71$\pm$0.69&&36.98$\pm$1.27 &48.07$\pm$1.66& 49.51$\pm$1.47\\
&&  \ding{51} Test&   82.55$\pm$0.86&76.33$\pm$0.95& 77.93$\pm$0.68 &78.42$\pm$0.64&&45.44$\pm$1.32 &60.23 $\pm$1.24&62.18$\pm$1.33\\
&&  \ding{51} Train+Test&86.69$\pm$0.12&84.06$\pm$0.21& \textbf{85.27$\pm$0.23} &86.06$\pm$0.20&&61.75$\pm$0.76 &\textbf{75.29$\pm$0.42}& \textbf{77.25$\pm$0.27}\\
\cline{2-11} } 
&Advers.& \ding{55}&69.32$\pm$1.61&61.73$\pm$1.12& 68.54$\pm$0.68 &68.00$\pm$0.31&&36.95$\pm$0.97 &50.21$\pm$0.55& 49.73$\pm$0.98
\\
\hline
\end{tabular}
}
\vspace{0.5pt}
\caption{Comparing  rotational  invariance  using  training  schemes  vs. orbit mapping for CIFAR10 classification using \textit{i)~Linear network ii)~5-layer Convnet iii)~Resnet18}. Shown are the mean clean accuracy and the average and worst case accuracies when test images are rotated in steps of 1 degree. The mean and standard deviation values over 5 runs are reported.}
\label{tab:cifar10_supple}
\end{table*}
\subsection{Invariance to continous image rotations}
We further investigate the effect of  discretization using different interpolation schemes for rotation on higher resolution on the CUB-200 dataset (trained at 224x224 resolution) fine-tuned using Resnet-50. Tab.~\ref{tab:rob.train_reg_cub.} shows the results of different training schemes with and without our orbit mapping (\textit{OM}) obtained when using different interpolation schemes for rotation. Besides standard training (\textit{Std.}), we use rotation augmentation (\textit{RA}), and the adversarial training and regularization from \cite{engstrom2019exploring,yang2019invariance}. Even for this higher resolution dataset, the worst-case accuracies between different types of interpolation may differ by more than $15\%$. In particular, adversarial training with bi-linear interpolation is still more vulnerable to image rotations with nearest neighbor interpolation. While our approach is also affected by the interpolation effects, the vulnerability to nearest neighbor interpolation is ameliorated when using rotation augmentation. 

\emph{Effect of Network architecture:}\hspace{.3cm} To investigate the effectiveness of our approach, we experiment three different network architectures:\textit{~i)~a linear network, ii)~a 5-layer convnet ii)~a Resnet18.} We compare the performance of our orbit mapping approach with training schemes, i.e. augmentation and adversarial training  for rotational invariance in Tab.~\ref{tab:cifar10_supple}. For all the three architectures considered,  our orbit mapping  together with rotation augmentation consistently results in  the most accurate predictions in the worst case.

\subsection{Pointcloud classification}
We add an additional point cloud experiment in Table~\ref{tab:3d_invariances}, where we show that in addition to being separately invariant to either scaling or rotation, the same approach can also work to create a model robust to both scaling and rotation. In our setup, we apply un-scaling before the un-rotation via PCA alignment.

\begin{table}
\small
\textbf{Simultaneous invariance to scaling and rotations:}\\
\begin{tabular}{c c | c| l | ll | ll | ll}
\multicolumn{2}{c|}{Augmentation} & Orbit & Clean & \multicolumn{2}{c|}{Scaling} &  \multicolumn{2}{|c|}{Rotation}\\
Random Scaling & Random Rotations &  & & Avg. & Worst & Avg. & Worst \\
\hline
$[0.1, 1000]$ & \ding{51} & All & 73.44 & 73.44 & 73.44 & 73.44 & 73.44 \\
$[0.8, 1.25]$ & \ding{51} & All & 73.48 & 73.48 & 73.48 & 73.48 & 73.48 \\
\end{tabular} 
\caption{Invariances in 3D pointcloud classification. Experiments with PointNet trained for classification of modelnet40 data, with data augmentation and analytical inclusion of each invariance. Shown is class accuracy of the final iterate on modelnet40 for the validation dataset \textit{(Clean)}, and  class accuracy in average- \textit{(Avg.)} and worst-case \textit{(Worst)} for scaled (left) and rotated (right) evaluations of each pointcloud in the validation set. In this experiment the model is invariant to \textit{both} scaling and translations. \label{tab:3d_invariances}}
\end{table}

\end{document}